\documentclass[10pt,twocolumn,letterpaper]{article}

\usepackage{cvpr}
\usepackage[T1]{fontenc}
\usepackage{times}
\usepackage{epsfig}
\usepackage{graphicx}
\usepackage{caption}
\usepackage{subfigure}
\usepackage{amsmath}
\usepackage{amssymb}
\usepackage{enumitem}

\usepackage[breaklinks=true,bookmarks=false]{hyperref}

\cvprfinalcopy 

\def\cvprPaperID{10165} 

\begin{document}

\title{Learning Invariant Representation for Unsupervised Image Restoration}

\author{
	Wenchao Du, Hu Chen$^{\dagger}$, Hongyu Yang\\
	College of Computer Science, Sichuan University, Chengdu 610065, China\\
	{\tt\small Wenchaodu.scu@gmail.com, huchen@scu.edu.cn, yanghongyu@scu.edu.cn}
}

\maketitle
\begin{abstract}
	\itshape 
	Recently, cross domain transfer has been applied for unsupervised image restoration tasks. However, directly applying existing frameworks would lead to domain-shift problems in translated images due to lack of effective supervision. Instead, we propose an unsupervised learning method that explicitly learns invariant presentation from noisy data and reconstructs clear observations. To do so, we introduce discrete disentangling representation and adversarial domain adaption into general domain transfer framework, aided by extra self-supervised modules including background and semantic consistency constraints, learning robust representation under dual domain constraints, such as feature and image domains. Experiments on synthetic and real noise removal tasks show the proposed method achieves comparable performance with other state-of-the-art supervised and unsupervised methods, while having faster and stable convergence than other domain adaption methods. \href{https://github.com/Wenchao-Du/LIR-for-Unsupervised-IR}{Code has been released}.
\end{abstract}

\section{Introduction}

Image restoration (IR) attempts to reconstruct clean signals from their corrupted observations, which is known to be an ill-posed inverse problem. By accommodating different types of corruption distributions, the same mathematical model applies to problems such as image denoising, super-resolution and deblurring. Recently, deep neural networks ({\itshape DNNs}) and generative adversarial networks ({\itshape GANs}) \cite{Goodfellow2014GenerativeAN} have shown their superior performance in various low-level vision tasks. Nonetheless, most of these methods need paired training data for specific tasks, which limits their generality, scalability and practicality in real-world multimedia applications. In addition, strong supervision may suffer from the overfitting training and lower generalization to real image corruption types.
\begin{figure}[t]
	\centering
	\subfigure[Input]{\includegraphics[width=0.24\linewidth]{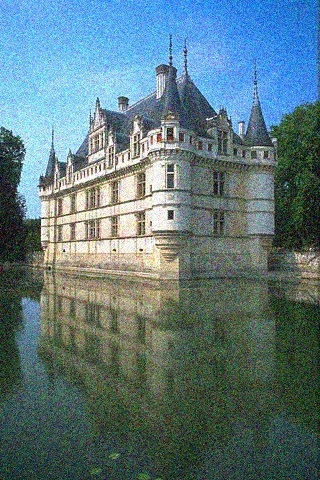}}
	\subfigure[{\itshape CycleGAN}]{\includegraphics[width=0.24\linewidth]{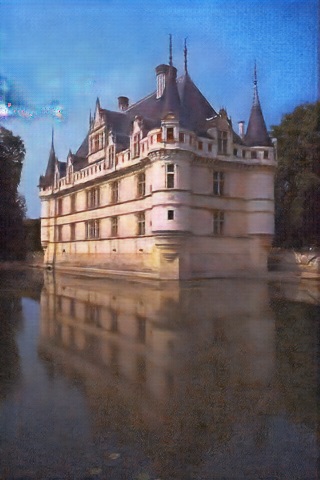}}
	\subfigure[{\itshape UNIT}]{\includegraphics[width=0.24\linewidth]{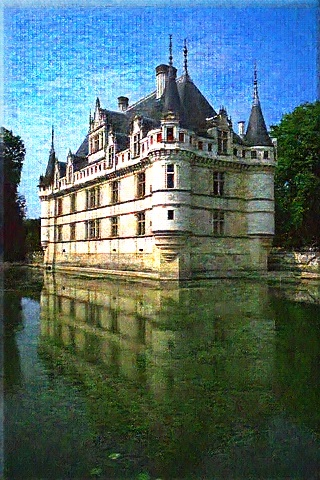}}
	\subfigure[{\itshape Ours}]{\includegraphics[width=0.24\linewidth]{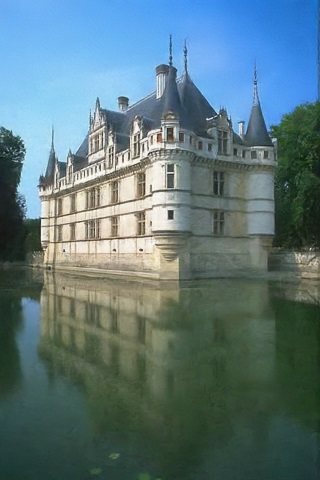}}
	\setlength{\abovecaptionskip}{-0.1cm}
	\caption{The typical results for Gaussian Noise. Our method has better ability on noise removal and texture preservation than other domain-transfer methods.}
	\label{fig:1}
\end{figure}

More recently, the domain transfer based unsupervised learning methods have attracted lots of attention due to the great progress \cite{Gatys2016ImageST,Lee2018DiverseIT,Liu2019STGANAU,Liu2017UnsupervisedIT,Zhu2017UnpairedIT} achieved in style transfer, attribute editing and image translation, \eg, {\itshape CycleGAN} \cite{Zhu2017UnpairedIT}, {\itshape UNIT} \cite{Liu2017UnsupervisedIT} and {\itshape DRIT} \cite{Lee2018DiverseIT}. Although these methods have been expanded to specific restoration tasks, they could not reconstruct the high-quality images due to losing finer details or inconsistency backgrounds, as shown in Fig. \ref{fig:1}. Different from {\itshape DNNs} based supervised models, which aim at learning a powerful mapping between the noisy and clean images. Directly applying existing domain-transfer methods is unsuitable for generalized image inverse problems due to the following reasons:
\begin{itemize}[leftmargin=*,topsep=2pt,itemsep=0pt]
	\item {\itshape Indistinct Domain Boundary.} Image translation aims to learn abstract shared-representations from unpaired data with clear domain characteristics, such as horse-to-zebra, day-to-night, etc. On the contrary, varying noise levels and complicated backgrounds blur domain boundaries between unpaired inputs.
	\item {\itshape Weak Representation.} Unsupervised domain-adaption methods extract high-level representations from unpaired data by shared-weight encoder and explicit target domain discriminator. For slight noisy signals, it is easy to cause domain shift problems in translated images and lead to low-quality reconstruction.
	\item {\itshape Poor Generalization.} Image translation learns a domain mapping from one-to-one image, which hardly captures the generalized semantic and texture representations. This also exacerbates the instability of GAN.
\end{itemize}

In order to address these problems, inspired by image sparse representation \cite{Mairal2008SparseRF} and domain adaption \cite{Ganin2014UnsupervisedDA,Ganin2015DomainAdversarialTO}, we attempt to learn invariant representation from unpaired samples via domain adaption and reconstruct clean images instead of relying on pure unsupervised domain transfer. Different from general image translation methods \cite{Lee2018DiverseIT,Liu2017UnsupervisedIT,Zhu2017UnpairedIT}, our goal is to learn robust intermediate representation free of noise (referred to as {\itshape Invariant Representation}) and reconstruct clean observations. Specifically, to achieve this goal, we factorize content and noise representations for corrupted images via disentangled learning; then a representation discriminator is utilized to align features to the expected distribution of clean domain. In addition, the extra self-supervised modules, including background and semantic consistency constraints, are used to supervise representation learning from image domains further.

In short, the main contributions of the paper could be summarized as follows: 1) Propose an unsupervised representation learning method for image restoration based on data-driven, which is easily expanded to other low-level vision tasks, such as super-resolution and deblurring. 2) Disentangle deep representation via dual domain constraints, \ie, feature and image domains. Extra self-supervised modules, including semantic meaning and background consistency modules, further improve the robustness of representations. 3) Build an unsupervised image restoration framework based on cross domain transfer with more effective training and faster convergence speed. To our knowledge, this is the first unsupervised representation learning approach that achieves competing results for processing synthetic and real noise removal with end-to-end training.
\section{Related Work}

\subsection{Single Image Restoraion}

{\bf \itshape Traditional Methods.} Classical methods, containing Total Variation \cite{Osher2005AnIR,Vese2004ImageDA}, {\itshape BM3D} \cite{Dabov2007ImageDB}, Non-local mean \cite{Buades2005ANA} and dictionary learning \cite{Chatterjee2009ClusteringBasedDW,Gu2015ConvolutionalSC}, have achieved good performance on general image restoration tasks, such as image denoising, super-resolution and deblurring. In addition, considering that image restoration is in general an ill-posed problem, some methods based on regularization are also proved effective \cite{Gu2014WeightedNN,Zoran2011FromLM}.

{\bf \itshape Deep Neural Networks.} Relying on powerful computer sources, data-driven DNN methods have achieved better performance than traditional methods in the past few years. Vincent \etal \cite{Vincent2008ExtractingAC} proposed stacked denoising auto-encoder for image restoration. Xie \etal \cite{Xie2012ImageDA} combined sparse coding and pre-trained DNN for image denoising and inpainting. Mao \etal \cite{Mao2016ImageRU} proposed {\itshape RedNet} with symmetric skip connections for noise removal and super-resolution. Zhang \etal \cite{Zhang2017BeyondAG} introduced residual learning for Gaussian noise removal. In general, {\itshape DNNs}-based methods could realize superior results on synthetic noise removal via effective supervised training, but it is unsuitable for real-world applications.

\subsection{Unsupervised Learning for IR}

{\bf \itshape Learning from noisy observations.} One interesting direction for unsupervised IR is directly recovering clean signals from noisy observations. Dmitry \etal \cite{Ulyanov2017DeepIP} proposed deep image prior ({\itshape DIP}) for IR, which requires suitable networks and interrupts its training process based on low-level statistical prior. That is usually unpredictable for different samples. Via zero-mean noise distribution prior, Noise2Noise ({\itshape N2N}) \cite{Lehtinen2018Noise2NoiseLI} directly learns reconstruction between two images with independent noise sampling. That is unsuitable for noise removal in real-world, \eg, medical image denoising. To alleviate this problem, {\itshape Noise2Void} \cite{Krull2018Noise2VoidL} predicted a pixel from its surroundings by learning a blind-spot network for corrupted images. Similar to {\itshape Noise2Self} \cite{Batson2019Noise2SelfBD}, this method reduces the training efficiency, but also decreases the denoising performance.

{\bf \itshape Image Domain Transfer.} Another direction solves image restoration by domain transfer, which aims to learn one2one mapping from one domain to another and output image to lie on the manifold of clean image. Previous works, \eg, {\itshape CycleGAN} \cite{Zhu2017UnpairedIT}, {\itshape DualGAN} \cite{Yi2017DualGANUD} and {\itshape BicycleGAN} \cite{Zhu2017TowardMI} have shown great capacity in image translation. Expanding works, containing {\itshape CouplesGAN} \cite{Liu2016CoupledGA}, {\itshape UNIT} \cite{Liu2017UnsupervisedIT} and {\itshape DRIT} \cite{Lee2018DiverseIT} learn shared-latent representation for diverse image translation. Along this way, Yuan \etal \cite{Yuan2018UnsupervisedIS} proposed a nested {\itshape CycleGAN} to solve the unsupervised image super-resolution. Expanding {\itshape DRIT}, Lu \etal \cite{Lu2019UnsupervisedDD} decoupled image content domain and blur domain to solve image deblurring, referred to as {\itshape DRNet}. However, these methods aim to learn stronger domain generators, they require obvious domain boundary and complicated network structure.
\begin{figure*}[t]
	\centering
	\subfigure[]{\includegraphics[width=0.33\linewidth, height=2in]{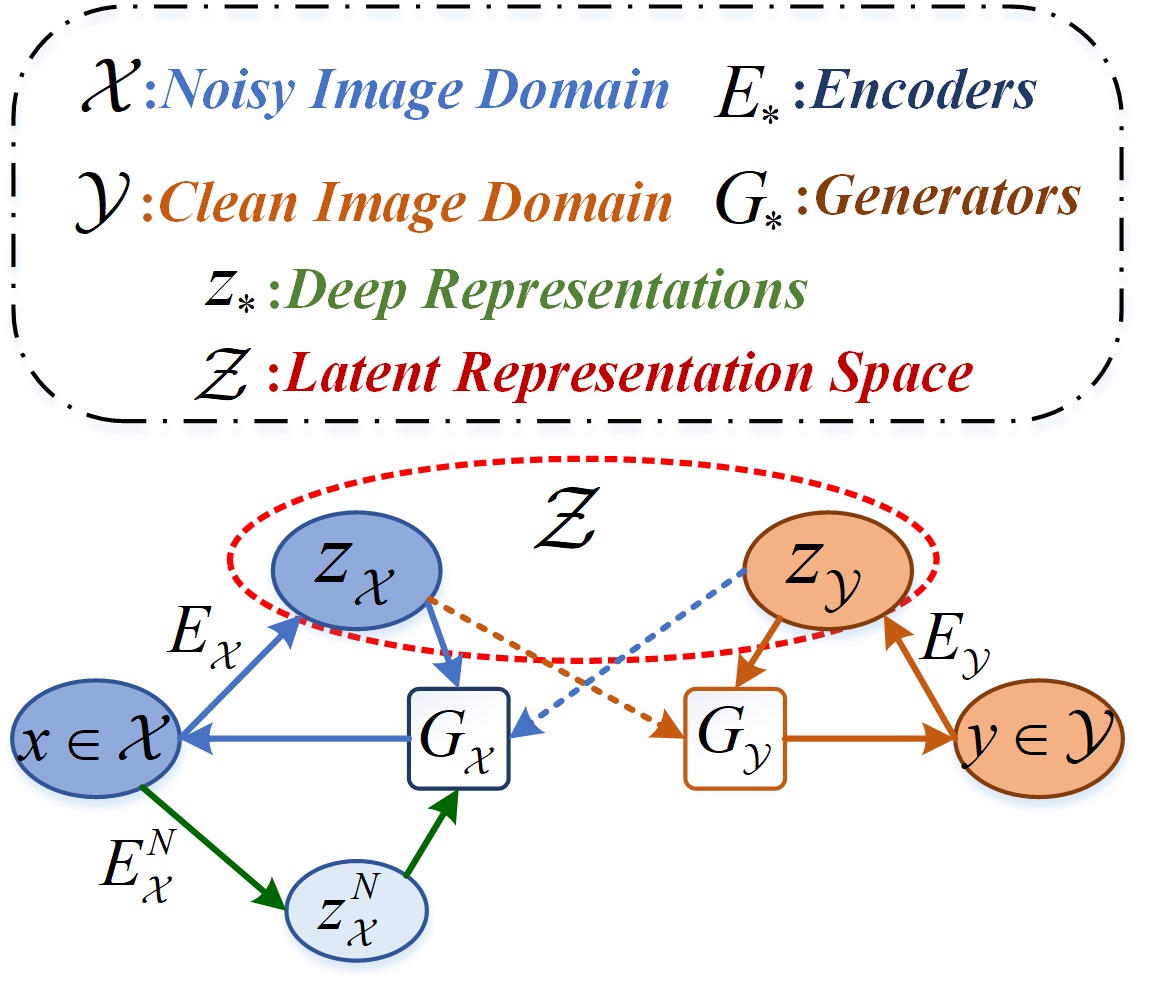}}
	\subfigure[]{\includegraphics[width=0.66\linewidth, height=2in]{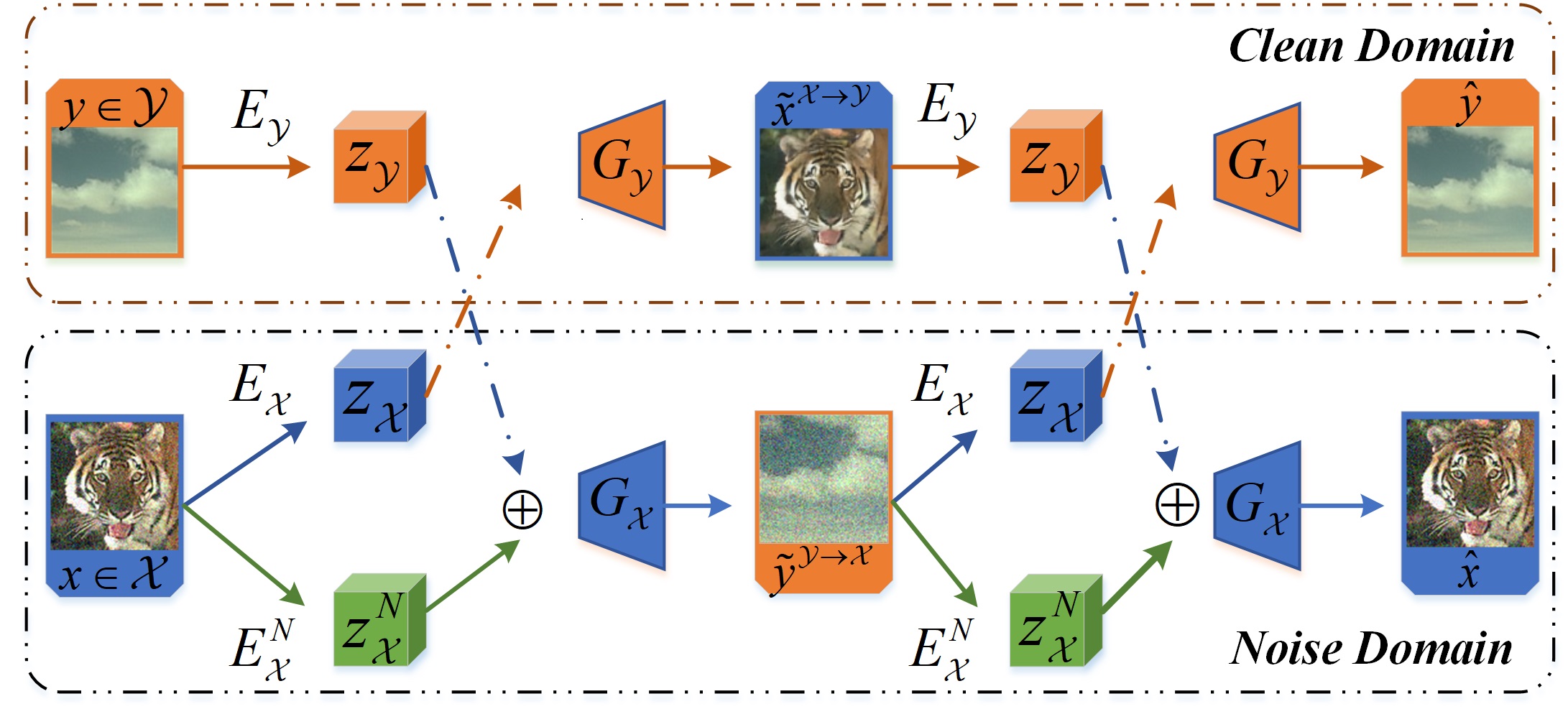}}
	\setlength{\abovecaptionskip}{-0.1cm}
	\caption{Method Overview. (a) The latent space assumption. Proposed method aims to learn invariant representations from inputs and align them via adversarial domain adaption. (b) Our method is injected into general domain-transfer framework. Extra self-supervised modules are introduced to learn more robust representations.}
	\label{fig:2}
\end{figure*}

\section{The Proposed Method}

Our goal is to learn abstract intermediate representations from noise inputs and reconstruct clear observations. In a certain way, unsupervised IR could be viewed as a specific domain transfer problem, \ie, from noise domain to clean domain. Therefore, the method is injected into the general domain transfer architecture, as shown in Fig. \ref{fig:2}.

In supervised domain transfer, we are given samples  $(x,y)$ drawn from a joint distribution ${P_{\mathcal{X},\mathcal{Y}}}(x,y)$, where $\mathcal{X}$ and $\mathcal{Y}$ are two image domains. For unsupervised domain translation, samples $(x,y)$ are drawn from the marginal distributions $P_{\mathcal{X}}(x)$ and $P_{\mathcal{Y}}(y)$. In order to infer the joint distribution from the marginal samples, a shared-latent space assumption is proposed that there exists a shared latent code $z$ in a shared-latent space $\mathcal{Z}$, so that we can recover both images from this code. Given samples $(x,y)$ from the joint distribution, this process is presented by
\begin{equation}
z=E_{\mathcal{X}}(x)=E_{\mathcal{Y}}(y)
\label{eq:1}
\end{equation}
\begin{equation}
x=G_{\mathcal{X}}(z), y=G_{\mathcal{Y}}(z)
\label{eq:2}
\end{equation}

A key step is how to implement this shared-latent space assumption. To do so, an effective strategy is sharing high-level representation by shared-weight encoder, which samples the features from the unified distribution. However, it is unsuitable for IR that latent representation only contains semantic meanings, which leads to domain shift in recovered images, \eg, blurred details and inconsistent backgrounds. Therefore, we attempt to learn more generalized representations containing richer texture and semantic features from inputs, \ie, invariant representations. To achieve it, adversarial domain adaption based discrete representation learning and self-supervised constraint modules are introduced into our method. Details are described in the subsections.
\subsection{Discrete Representation Learning}

Discrete representation aims to compute the latent code $z$ from inputs, where $z$ contains texture and semantic information as much as possible. To do so, we use two auto-encoders to model $\{E_{\mathcal{X}},G_{\mathcal{X}}\}$ and $\{E_{\mathcal{Y}},G_{\mathcal{Y}}\}$ separately. Given any unpaired samples $(x,y)$, where $x\in\mathcal{X}$ and $y\in\mathcal{Y}$ separately denote noise and clean sample from different domains, Eq. \ref{eq:1} is reformulated as ${z_{\mathcal{X}}}={E_{\mathcal{X}}}(x)$ and ${z_{\mathcal{Y}}}={E_{\mathcal{Y}}}(y)$. Further, IR could be represented as ${F^{\mathcal{X}\rightarrow \mathcal{Y}}}(x)={G_{\mathcal{Y}}}(z_{\mathcal{X}})$. However, considering noise always adheres to high-frequency signals, directly reconstructing clean images is difficult due to varying noise levels and types, which requires powerful domain generator and discriminator. Therefore, we introduce the disentangling representation into our architecture.

{\bf \itshape Disentangling Representation.} For noise sample $x$, an extra noise encoder $E^{N}_{\mathcal{X}}$ is used to model varying noisy levels and types. The self-reconstruction is formulated by $x={G_{\mathcal{X}}}({z_{\mathcal{X}}, z^{N}_{\mathcal{X}}})$, where ${z_{\mathcal{X}}}=E_{\mathcal{X}}(x)$ and ${z^{N}_{\mathcal{X}}}={E^{N}_{\mathcal{X}}}(x)$. Assuming the latent codes $z_{\mathcal{X}}$ and $z_{\mathcal{Y}}$ obey same distribution in shared-space that $\{{z_{\mathcal{X}},{z_{\mathcal{Y}}}}\}\in \mathcal{Z}$, similar to image translation, unsupervised image restoration could be divided into two stages: forward translation and back reconstruction.

{\bf \itshape Forward Cross Translation.} We first extract the representations $\{{z_{\mathcal{X}},{z_{\mathcal{Y}}}}\}$ from $(x,y)$ and extra noise code $z^{N}_{\mathcal{X}}$. Restoration and degradation could be represented by
\begin{equation}
\tilde{x}^{\mathcal{X}\rightarrow \mathcal{Y}}={G_{\mathcal{Y}}}(z_{\mathcal{X}})
\end{equation}
\begin{equation}
\tilde{y}^{\mathcal{Y}\rightarrow\mathcal{X}}={G_{\mathcal{X}}}(z_{\mathcal{Y}}\oplus z^{N}_{\mathcal{X}})
\end{equation}
where $\tilde {x}^{\mathcal{X}\rightarrow \mathcal{Y}}$ represents the recovered clean sample, $\tilde {y}^{\mathcal{Y}\rightarrow \mathcal{X}}$ denotes the degraded noise sample. $\oplus$ represents channel-wise concatenation operation. $G_{\mathcal{X}}$ and $G_{\mathcal{Y}}$ are viewed as specific domain generators.

{\bf \itshape Backward Cross Reconstruction.} After performing the first translation, reconstruction could be achieved by swapping the inputs $\tilde {x}^{\mathcal{X}\rightarrow \mathcal{Y}}$ and $\tilde {y}^{\mathcal{Y}\rightarrow \mathcal{X}}$ that:
\begin{equation}
\hat {x}=G_{\mathcal{X}}(E_{\mathcal{Y}}(\tilde {x}^{\mathcal{X}\rightarrow \mathcal{Y}})\oplus E^{N}_{\mathcal{X}}(\tilde {y}^{\mathcal{Y}\rightarrow \mathcal{X}}))
\end{equation}
\begin{equation}
\hat{y}=G_{\mathcal{Y}}(E_{\mathcal{X}}(\tilde{y}^{\mathcal{Y}\rightarrow \mathcal{X}}))
\end{equation}
where $\hat{x}$ and $\hat{y}$ denote reconstructed inputs. To enforce this constraint, we add the cross-cycle consistency loss $\mathcal {L}^{CC}$ for $\mathcal{X}$ and $\mathcal{Y}$ domains:
\begin{equation}
\begin{split}
\mathcal{L}^{CC}_{\mathcal{X}}(&G_{\mathcal{X}},G_{\mathcal{Y}},E_{\mathcal{X}},E_{\mathcal{Y}}, E^{N}_{\mathcal{X}})=\\&\mathbb{E}_{\mathcal{X}}\left[\Vert G_{\mathcal{X}}(E_{\mathcal{Y}}(\tilde{x}^{\mathcal{X}\rightarrow \mathcal{Y}})\oplus E^{N}_{\mathcal{X}}(\tilde{y}^{\mathcal{Y}\rightarrow\mathcal{X}}))-x\Vert_{1} \right]
\end{split}
\end{equation}
\begin{equation}
\begin{split}
\mathcal{L}^{CC}_{\mathcal{Y}}(G_{\mathcal{X}},G_{\mathcal{Y}},E_{\mathcal{X}},&E_{\mathcal{Y}}, E^{N}_{\mathcal{X}})=\\&\mathbb{E}_{\mathcal{Y}}\left[\Vert G_{\mathcal{Y}}(E_{\mathcal{X}}(\tilde{y}^{\mathcal{Y}\rightarrow \mathcal{X}})-y\Vert_{1} \right]
\end{split}
\end{equation}

{\bf \itshape Adversarial Domain Adaption.} Another factor is how to embed latent representations $z_{\mathcal{X}}$ and $z_{\mathcal{Y}}$ into shared space. Inspired by unsupervised domain adaption, we implement it by adversarial learning instead of shared-weight encoder. Our goal is to facilitate representations from inputs obeying the similar distribution while preserving richer texture and semantic information of inputs. Therefore, a representation discriminator $D_{\mathcal{R}}$ is utilized in our architecture. We express this feature adversarial loss $\mathcal{L}^{\mathcal{R}}_{adv}$ as
\begin{equation}
\begin{split}
\mathcal{L}^{\mathcal{R}}_{adv}(&E_{\mathcal{X}},E_{\mathcal{Y}},D_{\mathcal{R}})=\\
\mathbb{E}_{\mathcal{X}}&\left[\frac{1}{2}\*\log D_{\mathcal{R}}(z_{\mathcal{X}})+\frac{1}{2}\*\log(1-D_{\mathcal{R}}(z_{\mathcal{X}}))\right]+\\
&\mathbb{E}_{\mathcal{Y}}\left[\frac{1}{2}\*\log D_{\mathcal{R}}(z_{\mathcal{Y}})+\frac{1}{2}\*\log(1-D_{\mathcal{R}}(z_{\mathcal{Y}}))\right]
\end{split}
\end{equation}
\begin{figure}[t]
	\centering
		\includegraphics[width=1.0\linewidth]{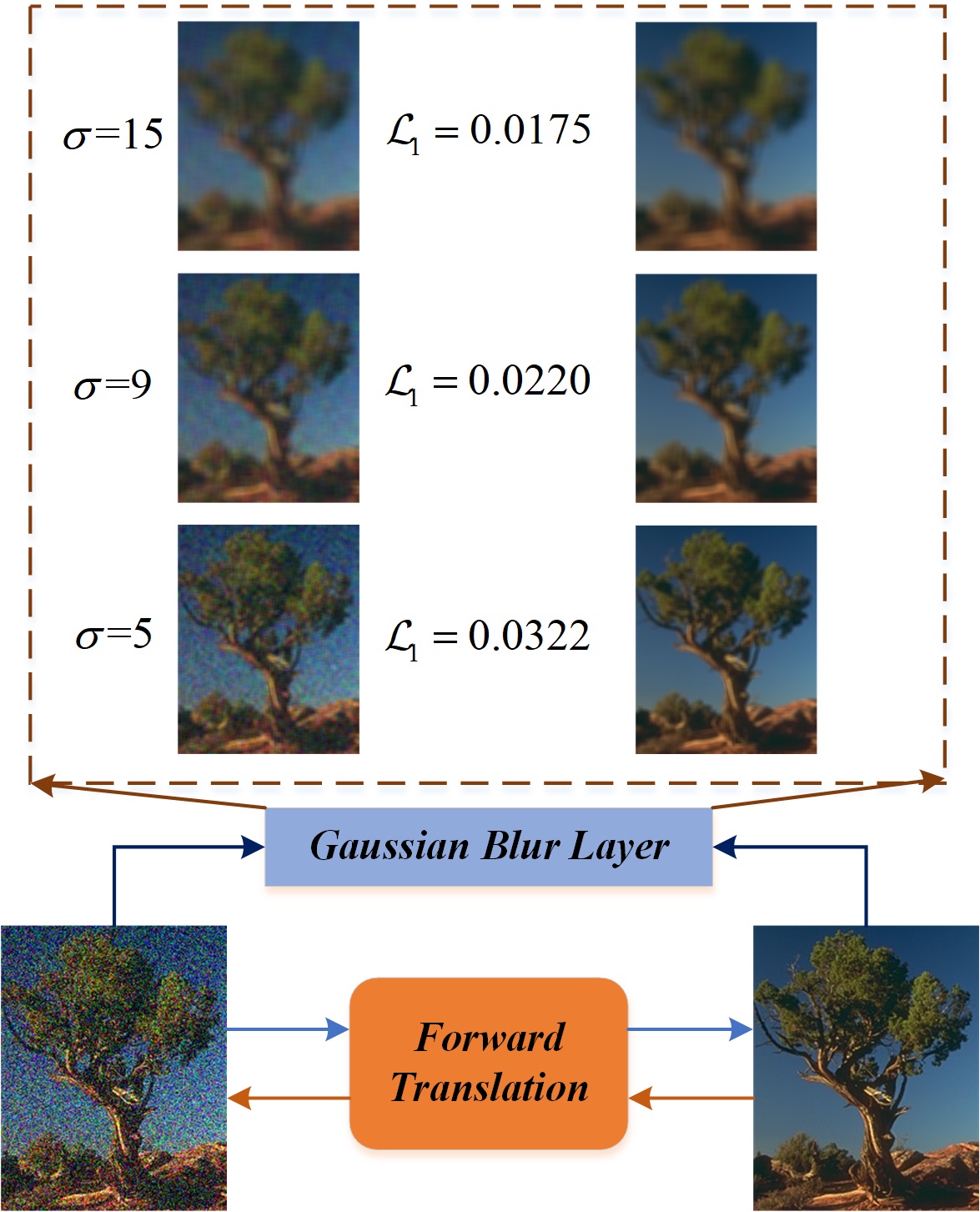}
		\caption{{\itshape Background Consistency Module (BCM)}. {\itshape BCM}
			hierarchically uses $\mathcal{L}_{1}$ loss at different Gaussian-Blur levels to ensure the inputs and outputs have consistency background.}
		\label{fig:3}
\end{figure}

\begin{figure*}[ht]
	\begin{minipage}{1.0\linewidth}
		\centering
			\includegraphics[width=1.0\linewidth]{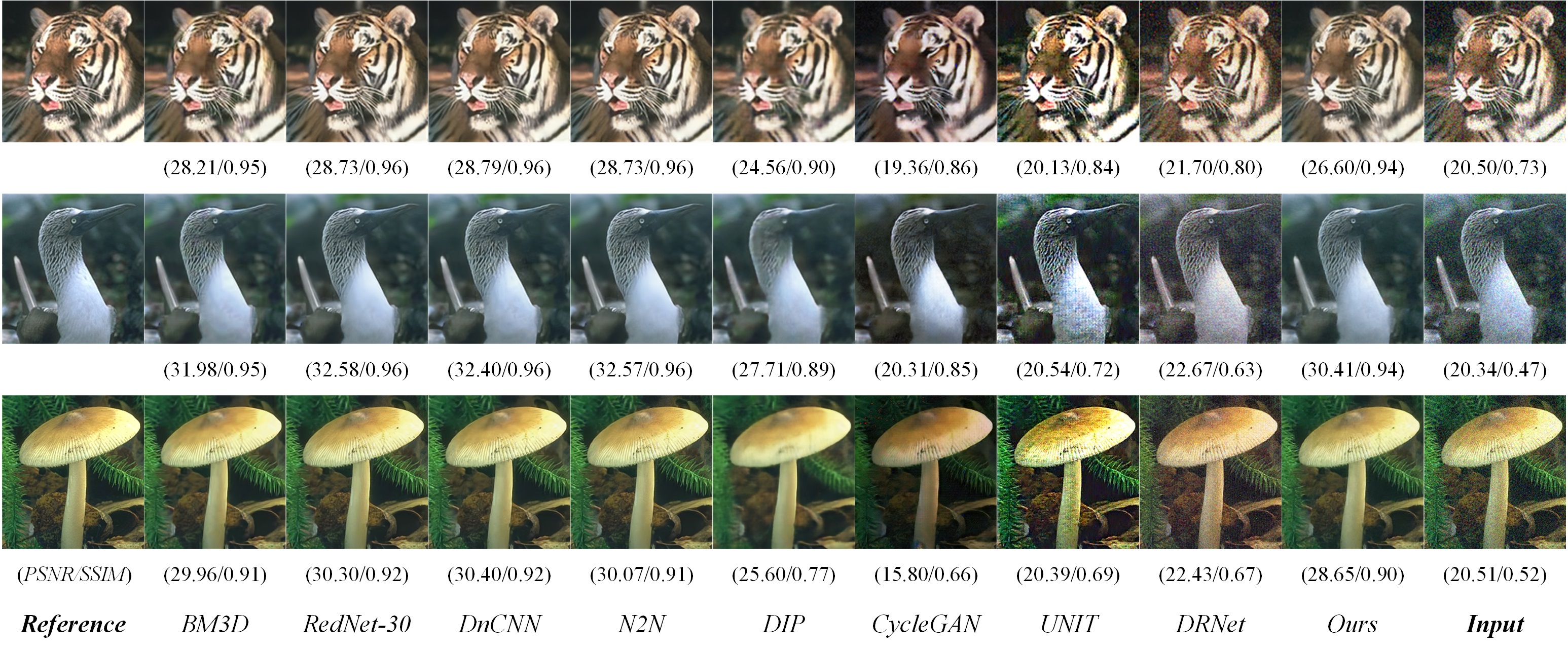}
			\setlength{\abovecaptionskip}{-0.4cm}
			\setlength{\belowcaptionskip}{0.1cm}
			\caption{The example results for Gaussian noise on BSD-68. Zooming in for better visualization.}
			\label{fig:4}
	\end{minipage}
	\begin{minipage}{1.0\linewidth}
		\centering
			\footnotesize
			\renewcommand\tabcolsep{5.0pt}
			\begin{tabular}{c|ccccccccc}
				\hline
				Methods & {\itshape BM3D}\cite{Dabov2007ImageDB} & {\itshape RedNet-30}\cite{Mao2016ImageRU} & {\itshape DnCNN}\cite{Zhang2017BeyondAG} & {\itshape N2N}\cite{Lehtinen2018Noise2NoiseLI} & {\itshape DIP}\cite{Ulyanov2017DeepIP} & {\itshape CycleGAN}\cite{Zhu2017UnpairedIT} & {\itshape UNIT}\cite{Liu2017UnsupervisedIT} & {\itshape DRNet}\cite{Lu2019UnsupervisedDD} & {\itshape Ours}\\
				\hline
				\multicolumn{1}{c|}{} & \multicolumn{9}{c}{PSNR$\left(mean\pm std\right)$}\\
				\hline
				$\sigma=25$&
				30.18$ \pm $2.07 &30.19$ \pm $2.07&30.70$ \pm $2.04&30.21$ \pm $2.19&26.48$ \pm $3.14&19.08$ \pm $2.27&20.21$ \pm $1.45&21.06$ \pm $2.23&{\bf 29.02$ \pm $1.93}\\
				$\sigma=35$&
				28.09$ \pm $2.17 &28.27$ \pm $2.30&28.75$ \pm $2.10&28.28$ \pm $2.29&26.06$ \pm $2.78&16.77$ \pm $1.63&18.96$ \pm $1.29&19.10$ \pm $1.70&{\bf 27.58$ \pm $1.98}\\
				$\sigma=50$&
				25.87$ \pm $2.31 &25.22$ \pm $2.84&26.54$ \pm $2.15&25.85$ \pm $2.58&24.80$ \pm $2.25&16.68$ \pm $2.35&17.10$ \pm $1.08&16.78$ \pm $1.22&{\bf 24.69$ \pm $1.59}\\
				\hline
				\multicolumn{1}{c|}{} & \multicolumn{9}{c}{SSIM$\left(mean\pm std\right)$}\\
				\hline
				$\sigma=25$&
				0.921$ \pm $0.03 &0.918$ \pm $0.03&0.931$ \pm $0.02&0.919$ \pm $0.03&0.820$ \pm $0.09&0.808$ \pm $0.06&0.709$ \pm $0.08&0.626$ \pm $0.09&{\bf 0.917$ \pm $0.02}\\
				$\sigma=35$&
				0.883$ \pm $0.04 &0.885$ \pm $0.04&0.901$ \pm $0.03&0.886$ \pm $0.04&0.817$ \pm $0.07&0.731$ \pm $0.07&0.599$ \pm $0.10&0.505$ \pm $0.09&{\bf 0.887$ \pm $0.03}\\
				$\sigma=50$&
				0.830$ \pm $0.06 &0.827$ \pm $0.06&0.857$ \pm $0.05&0.832$ \pm $0.06&0.786$ \pm $0.07&0.696$ \pm $0.06&0.459$ \pm $0.11&0.374$ \pm $0.08&{\bf 0.787$ \pm $0.04}\\
				\hline
			\end{tabular}
		\captionof{table}{Quantitative results for Gaussian noise reduction on BSD-68 dataset.}
		\label{table:1}
	\end{minipage}
\end{figure*}

\subsection{Self-Supervised Constraint}

Due to lack of effective supervised signals for translated images, only relying on feature domain discriminant constraints would lead to domain shift problems inevitably in generated images. To speed convergence while learning more robust representations, self-supervised modules including {\itshape Background Consistency Module (BCM)} and {\itshape Semantic Consistency Module (SCM)} are introduced to provide more reasonable and reliable supervision.

{\itshape BCM} aims to preserve the background consistency between the translated images and inputs. Similar strategies have been applied for self-supervised image reconstruction tasks \cite{Jin2018UnsupervisedSI,Nimisha2018UnsupervisedCD}. These methods use the gradient error to constrain reconstructed images by smoothing the input and output images with blur operators, \eg, Gaussian blur kernel and guided filtering \cite{He2013GuidedIF}. Different from them, a $\mathcal{L}_1$ loss is directly used for the recovered images instead of gradient error loss in our module, as shown in Fig. \ref{fig:3}, which is simple but effective to retain background consistency while recovering finer texture in our experiments. Specifically, a multi-scale Gaussian-Blur operator is used to obtain multi-scale features respectively. Therefore, a background consistency loss $\mathcal{L}_{BC}$ could be formulated as:
\begin{equation}
\mathcal{L}_{BC}=\sum_{\sigma=5,9,15}\lambda_{\sigma}\Vert B_{\sigma}(\chi)-B_{\sigma}(\tilde{\chi}) \Vert_1
\end{equation}
where $B_{\sigma}(\cdot)$ denotes the Gaussian-Blur operator with blur kernel $\sigma$, $\lambda_{\sigma}$ is the hyper-parameter to balance the errors at different Gaussian-Blur levels. $\chi$ and $\tilde{\chi}$ denote original input and the translated output, \ie, $\{x, \tilde{x}^{\mathcal{X}\rightarrow\mathcal{Y}}\}$ and $\{y, \tilde{y}^{\mathcal{Y}\rightarrow\mathcal{X}}\}$. Based on experimental attempts at image denoising, we set $\lambda_{\sigma}$ as $\{0.25,0.5,1.0 \}$ for $\sigma=\{5,9,15\}$ respectively.

In addition, inspired by perception loss \cite{Johnson2016PerceptualLF}, the feature from the deeper layers of the pre-trained model contain semantic meanings only, which are noiseless or with little noise. Therefore, different from the general feature loss, which aims to recover finer image texture details via similarities among shallow features, we only extract deeper features as semantic representations from the corrupted and recovered images to keep consistency, referred to as semantic consistency loss $\mathcal{L}_{SC}$. It could be formulated as
\begin{equation}
\mathcal{L}_{SC}=\Vert\phi_{l}(\chi)-\phi_{l}(\tilde{\chi}) \Vert^{2}_{2}
\end{equation}
where $\phi (\cdot)$ denotes the features from $l_{th}$ layer of the pre-trained model. In our experiments, we use the {\itshape conv5-1} layer of VGG-19 \cite{Simonyan2014VeryDC} pre-trained network on ImageNet.

\subsection{Jointly Optimizing}

Other than proposed cross-cycle consistency loss, representation adversarial loss and self-supervised loss, we also use other loss functions in our joint optimization.

{\bf \itshape Target Domain Adversarial Loss.} We impose domain adversarial loss $\mathcal{L}^{domain}_{adv}$, where $D_{\mathcal{X}}$ and $D_{\mathcal{Y}}$ attempt to discriminate the realness of generated images from each domain. For the noise domain, we define the adversarial loss $\mathcal{L}^{\mathcal{X}}_{adv}$ as
\begin{equation}
\begin{split}
&\mathcal{L}^{\mathcal{X}}_{adv}=\mathbb{E}_{x\thicksim P_{\mathcal{X}}(x)} \left[\log D_{\mathcal{X}}(x) \right]+\\
&\mathbb{E}_{\substack{y\thicksim P_{\mathcal{Y}}(y) \\ x\thicksim P_{\mathcal{X}}(x)}}\left[\log(1-D_{\mathcal{X}}(G_{\mathcal{X}}(E_{\mathcal{Y}}(y),E^{N}_{\mathcal{X}}(x)))) \right]
\end{split}
\end{equation}
Similarly, we define adversarial loss for clean image domain as
\begin{equation}
\begin{split}
\mathcal{L}^{\mathcal{Y}}_{adv}=&\mathbb{E}_{y\thicksim P_{\mathcal{Y}}(y)}\left[\log D_{\mathcal{Y}}(y)\right]+\\
&\mathbb{E}_{x\thicksim P_{\mathcal{X}}(x)}\left[\log(1-D_{\mathcal{Y}}(G_{\mathcal{Y}}(E_{\mathcal{X}}(x))))\right]
\end{split}
\end{equation}

{\bf \itshape Self-Reconstruction Loss.} In addition to the cross-cycle reconstruction, we also apply a self-reconstruction loss $\mathcal{L}^{Rec}$ to facilitate the training. This process is represented as $\hat{x}=G_{\mathcal{X}}(E_{\mathcal{X}}(x)\oplus E^{N}_{\mathcal{X}}(x))$ and $\hat{y}=G_{\mathcal{Y}}(E_{\mathcal{Y}}(y))$.

{\bf \itshape KL Loss.} In order to model the noise encoder branch, we add a KL divergence loss to regularize the distribution of the noise code $z^{N}_{\mathcal{X}}=E^{N}_{\mathcal{X}}(x)$ to be close to the normal distribution that $p(z^{N}_{\mathcal{X}}\thicksim N(0,1))$, where $D_{KL}=-\int p(z)\log(\frac{p(z)}{q(z)})dz $.

The full objective function of our method is summarized as follows:
\begin{equation}
\begin{split}
\min_{E_{\mathcal{X}},E^{N}_{\mathcal{X}},E_{\mathcal{Y}},G_{\mathcal{X}},G_{\mathcal{Y}}}\max_{D_{\mathcal{X}},D_{\mathcal{Y}},D_{\mathcal{R}}}=&\lambda_{\mathcal{R}}\mathcal{L}^{\mathcal{R}}_{adv}+ \\ 
\lambda_{adv}\mathcal{L}^{domain}_{adv}+\lambda_{CC}&\mathcal{L}^{CC}+\lambda_{rec}\mathcal{L}^{Rec}+\\
\lambda_{bc}\mathcal{L}^{BC}+&\lambda_{sc}\mathcal{L}^{SC}+\lambda_{KL}\mathcal{L}^{KL}
\end{split}
\end{equation}
where the hyper-parameters $\lambda_{\ast}$ control the importance of each term.

{\bf \itshape Restoration}: After learning, we only retain the cross encoder-generator network $\{E_{\mathcal{X}}, G_{\mathcal{Y}}\}$, $E_{\mathcal{X}}$ extracts the domain-invariant representation $z_{\mathcal{X}}$ from corrupted sample $x$, and $G_{\mathcal{Y}}$ recover the clean image $\tilde{x}^{\mathcal{X}\rightarrow \mathcal{Y}}$ from the $z_{\mathcal{X}}$ that $\tilde{x}^{\mathcal{X}\rightarrow \mathcal{Y}}=G_{\mathcal{Y}}(E_{\mathcal{X}}(x))$.

\begin{figure*}[ht]
	\centering
	\includegraphics[width=1.0\linewidth]{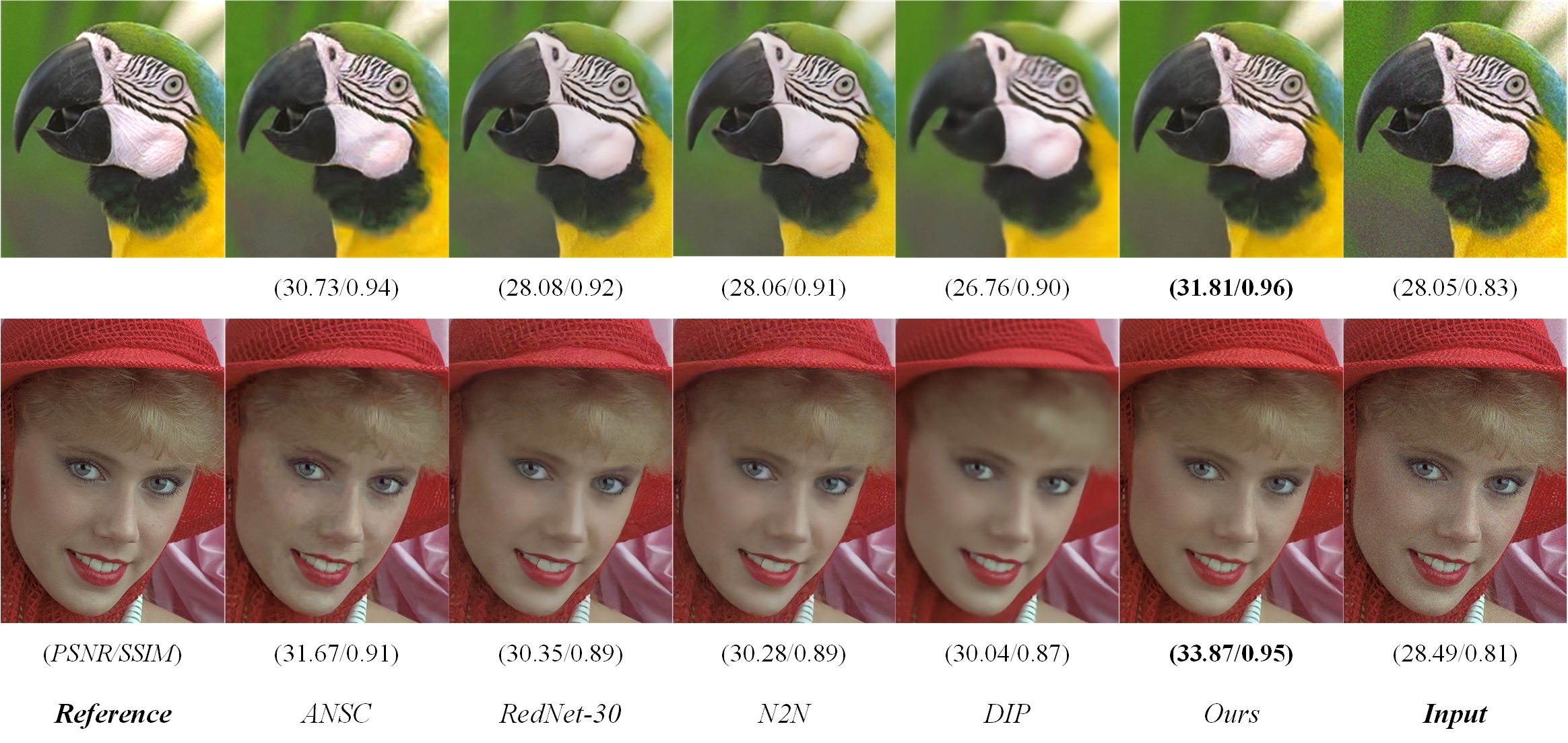}
	\setlength{\abovecaptionskip}{-0.4cm}
	\setlength{\belowcaptionskip}{-0.1cm}
	\caption{Sample results from Kodak dataset. Best detail visualization by zooming in.}
	\label{fig:5}
\end{figure*}

\section{Experiments}

In this section, we first give the implementation details of our method for classical image denoising. Traditional metrics, such as Peak-Signal-Noise-Rate (PSNR) and Structural Similarity (SSIM), are used for evaluation in experiments. Detailed results on synthetic and real noise removal tasks are shown with other state-of-the-art methods. For the synthetic noise removal, we start with general noise distributions including additive white Gaussian noise (AWGN) and Poisson noise. Two well-known datasets BSD68 \cite{Martin2001ADO} and Kodak are used to verify the performance of our method in denoising and texture restoration. Furthermore, the real noise images from the medical Low-Dose Computed Tomography (LDCT) dataset are used to evaluate the generalized capacity of the method. Extra ablation study is used to verify the effectiveness of the proposed framework.

\begin{table}[t]
	\centering
	\setlength{\tabcolsep}{1pt}
	\begin{tabular}{lcc}
		\hline
		Method & PSNR$\left(mean\pm std\right)$ & SSIM $\left(mean\pm std\right)$\\
		\hline
		{\itshape DIP}\cite{Ulyanov2017DeepIP} & $27.63\pm 2.66$ & $0.838\pm 0.07$\\
		{\itshape N2N}\cite{Lehtinen2018Noise2NoiseLI} & $28.39\pm 2.04$ & $0.893\pm 0.03$\\
		{\itshape ANSC}\cite{Mkitalo2011OptimalIO}& $30.68\pm 1.81$&$0.918\pm 0.02$\\
		{\itshape RedNet-30}\cite{Mao2016ImageRU}&$28.34\pm 2.07$& $0.893\pm 0.03$\\
		{\itshape Ours}&{\bf 32.37 $\pm$ 1.55}& {\bf 0.957 $\pm$ 0.01}\\
		\hline
	\end{tabular}
	\setlength{\belowcaptionskip}{-0.5cm}
	\caption{Quantitative results for Poisson noise.}
	\label{table:2}
\end{table}

\subsection{Implementation}

We follow the similar network architecture as the one used in \cite{Liu2017UnsupervisedIT}, the difference is we introduce an extra noise encoder branch and remove the shared-weight encoder. Representation discriminator is a full convolutional network structure, which stacks four convolutional layers with two strides and a global average pooling layer. Proposed framework is implemented with Pytorch \cite{Paszke2017AutomaticDI} and an Nvidia TITAN-V GPU is used in experiments. During the training, we use Adam \cite{Kingma2014AdamAM} to perform optimization and momentum is set to 0.9. The learning rate is initially set to 0.0001 and exponential decay over the 10K iterators. In all experiments, we randomly crop 64$\times$64 patches with batch size of 16 for training. Hyper-parameters are set to $\lambda_{\mathcal{R}}=\lambda^{domain}_{adv}=\lambda_{sc}=1$, $\lambda_{cc}=\lambda_{rec}=10$, $\lambda_{bc}=5$ and $\lambda_{KL}=0.01$.

\subsection{Synthetic Noise Removal}

We train the model with the images from the Pascal2007 \cite{Everingham2009ThePV} training set. Samples are randomly divided into two parts without coinciding. We add different noise-levels to each sample in part one, which is viewed as corrupted set, and another is clean set. Proposed method needs to estimate the magnitude of noise while removing it (“blind” image denoising). Some supervised and unsupervised based methods are selected to evaluate.

{\bf \itshape AWGN Removal.} We add the AWGN with zero mean and standard deviation   randomly generated with ranges from 5 to 50 for each training example, test on BSD68 with $\sigma=\{25,35,50\}$. The representative unsupervised methods, including {\itshape DIP} \cite{Ulyanov2017DeepIP}, {\itshape Noise2Noise} ({\itshape N2N}) \cite{Lehtinen2018Noise2NoiseLI}, {\itshape CycleGAN} \cite{Zhu2017UnpairedIT}, {\itshape UNIT} \cite{Liu2017UnsupervisedIT} and {\itshape DRNet} \cite{Lu2019UnsupervisedDD}, and supervised methods (\eg, {\itshape RedNet-30} \cite{Mao2016ImageRU} and {\itshape DnCNN} \cite{Zhang2017BeyondAG}), are selected to compare the performance on image denoising. Traditional {\itshape BM3D} is also included for evaluation. For {\itshape CycleGAN}, {\itshape UNIT} and {\itshape DRNet}, we retrain them with the same training data.

The visualized results from BSD68 dataset are given in Fig. \ref{fig:4}. Although all the methods show the ability for noise reduction, domain transfer based unsupervised methods, including {\itshape CycleGAN}, {\itshape UNIT} and {\itshape DRNet}, have obvious domain shift problems, \eg, inconsistent brightness and undesired artifacts, resulting in worse visual perception. {\itshape N2N} and {\itshape DIP} achieve higher PSNR and SSIM. However, {\itshape DIP} loses fine local details and leads to over-smoothness in the generated images. Depending on the zero-mean distribution prior, {\itshape N2N} achieves similar results with other supervised methods, such as {\itshape RedNet-30} and {\itshape DnCNN}. Our approach presents comparable performance on noise removal and texture preserving. Although the PSNR is slightly lower than other supervised methods’, our method achieves better visual consistency with natural images. Quantitative results for BSD68 are given in Table. \ref{table:1}. The proposed method shows stronger ability to blind image denoising.

{\bf \itshape Poisson Noise Removal.} For corrupted samples, we randomly generate the noise data from Scikit-image library \cite{Walt2014scikitimageIP}, which generates independent Poisson noise by the number of unique values in the given samples, and test on Kodak\footnote{\url{http://r0k.us/graphics/kodak/}} dataset. Some representative methods, including {\itshape DIP}, {\itshape N2N}, {\itshape ANSC} \cite{Mkitalo2011OptimalIO} and {\itshape RedNet-30}, are selected in our evaluations.

Comprehensive results are shown in Fig. \ref{fig:5} and Table. \ref{table:2}. {\itshape DIP} tends to generate more blurred results. The traditional {\itshape ANSC} method first transforms the Poisson noise into Gaussian (Anscombe transform), then applies the {\itshape BM3D} to remove noise, and finally inverts the transform, achieving higher PSNR and SSIM. Considering the different way of generating Poisson noise, the published {\itshape RedNet-30} and {\itshape N2N} models don’t achieve the best results. Our method achieves the highest PSNR and SSIM. In addition, visualized results also show that for slight noise signals, the proposed framework has better generalized capacity to remove noise while restoring finer details.

\begin{figure}[ht]
	\begin{minipage}{1.0\linewidth}
		\centering
		\includegraphics[width=1.0\linewidth]{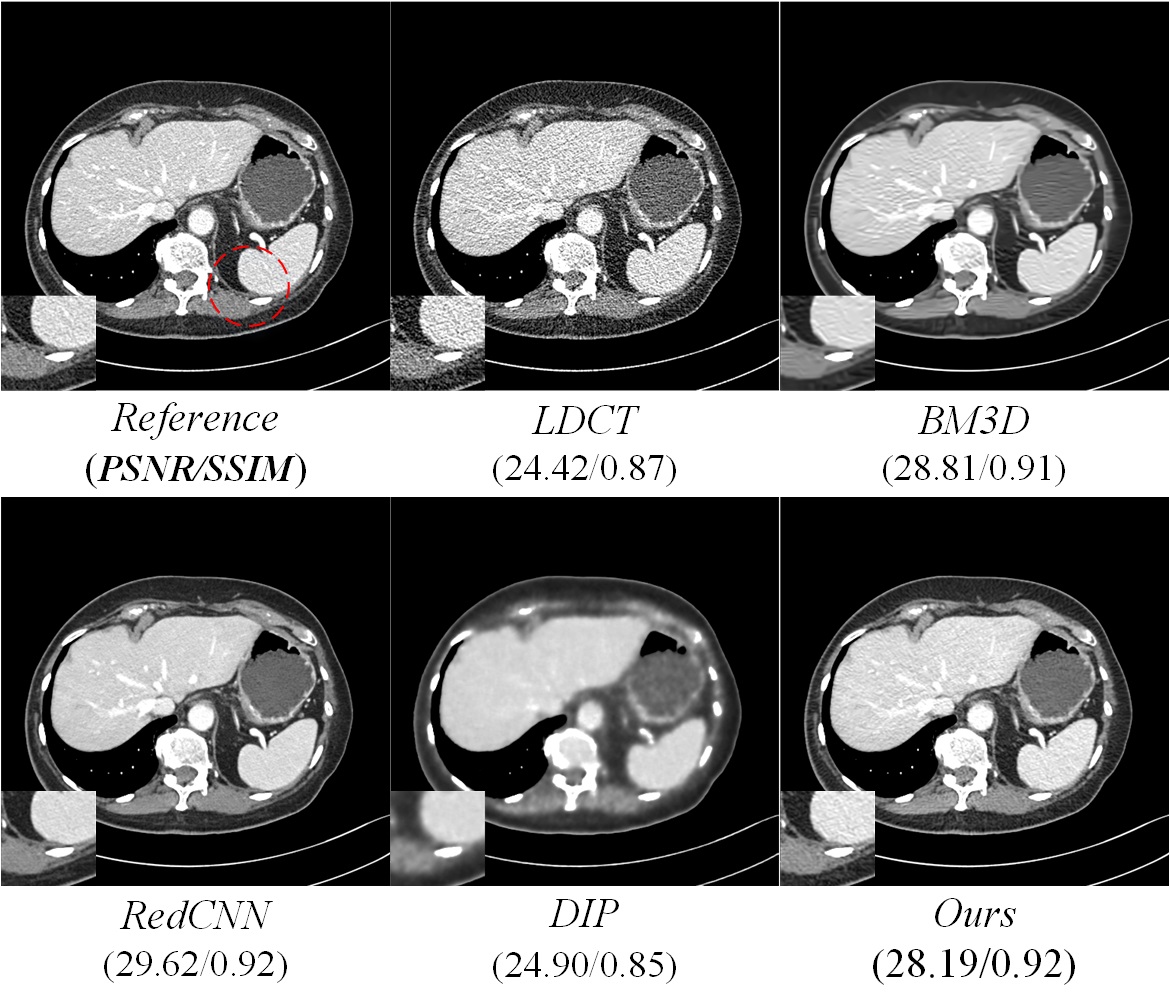}
		\setlength{\abovecaptionskip}{-0.3cm}
		\setlength{\belowcaptionskip}{0.1cm}
		\caption{LDCT Reconstruction. The display window is $\left[160,240\right]$HU. The red circle denotes ROI area.}
		\label{fig:6}
	\end{minipage}
	\begin{minipage}{1.0\linewidth}
		\centering
		\setlength{\tabcolsep}{10pt}
		\begin{tabular}{lcc}
			\hline
			Methods & PSNR & SSIM \\
			\hline
			{\itshape LDCT} & 36.3616 & 0.9423\\
			{\itshape BM3D}\cite{Dabov2007ImageDB} & 40.6941 & 0.9755\\
			{\itshape RedCNN}\cite{Chen2017LowDoseCW} & 41.8799 & 0.9846\\
			{\itshape DIP}\cite{Ulyanov2017DeepIP} & 36.2047 & 0.9500\\
			{\itshape Ours} & 40.5857 & 0.9811\\
			\hline
		\end{tabular}
		\setlength{\belowcaptionskip}{-0.1cm}
		\captionof{table}{ Quantitative results on Mayo dataset.}
		\label{table:3}
	\end{minipage}
\end{figure}

\begin{figure*}[ht]
	\centering
		\includegraphics[width=1.0\linewidth]{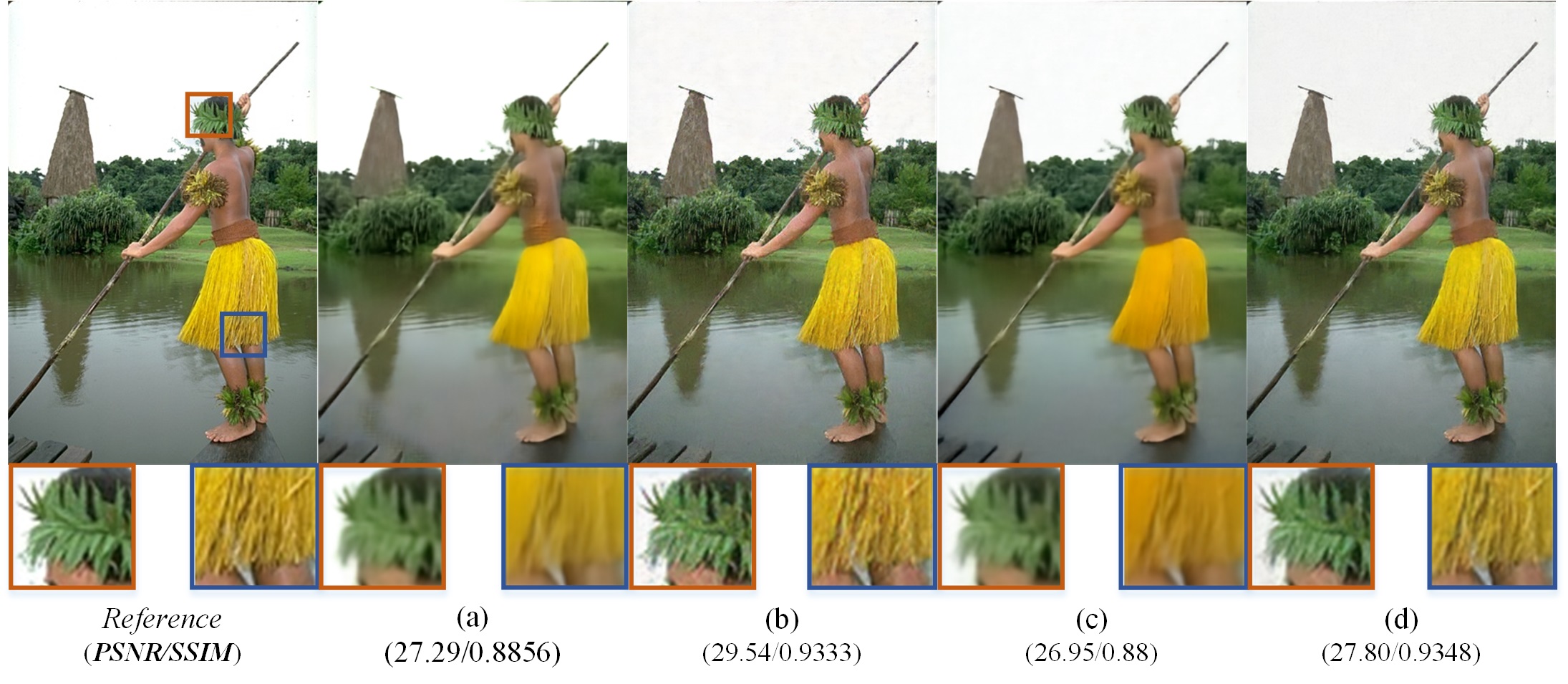}
		\setlength{\abovecaptionskip}{-0.2cm}
		\caption{The visualized results for each variants. (a) Without $E^{N}_{\mathcal{X}}$. (b) Without $D_{\mathcal{R}}$. (c) Removing {\itshape BGM} (d) Full model.}
		\label{fig:7}
\end{figure*}

\begin{figure}[t]
	\begin{minipage}{1.0\linewidth}
		\centering
		\begin{tabular}{ccc}
			\hline
			Variants & PSNR$\left(mean\pm std\right)$ & SSIM $\left(mean\pm std\right)$\\
			\hline
			$\left(a\right)$ & $25.997\pm 1.50$ & $0.828\pm 0.07$\\
			$\left(b\right)$ & $29.452\pm 1.71$ & $0.913\pm 0.02$\\
			$\left(c\right)$ & $25.220\pm 1.62$ & $0.817\pm 0.08$\\
			$\left(d\right)$ & {\bf 29.022 $\pm$ 1.93} & {\bf 0.917 $\pm$ 0.02}\\
			\hline
		\end{tabular}
		\captionof{table}{ Quantitative results for Gaussian noise with $\sigma=25$ on BSD-68.}
		\label{table:4}
	\end{minipage}
		\begin{minipage}{1.0\linewidth}
			\centering
			\subfigure[]{\includegraphics[width=0.46\linewidth]{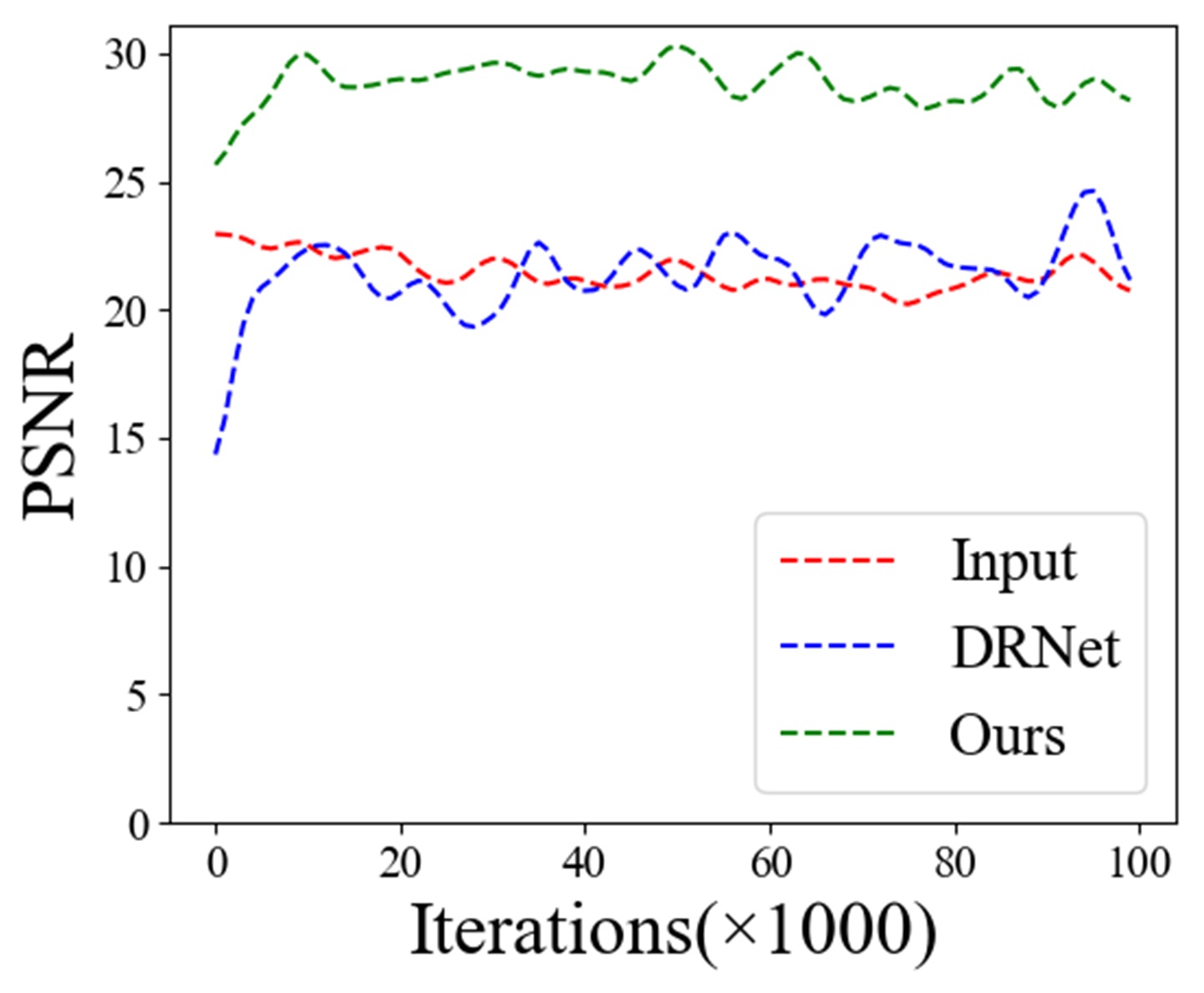}}
			\subfigure[]{\includegraphics[width=0.46\linewidth]{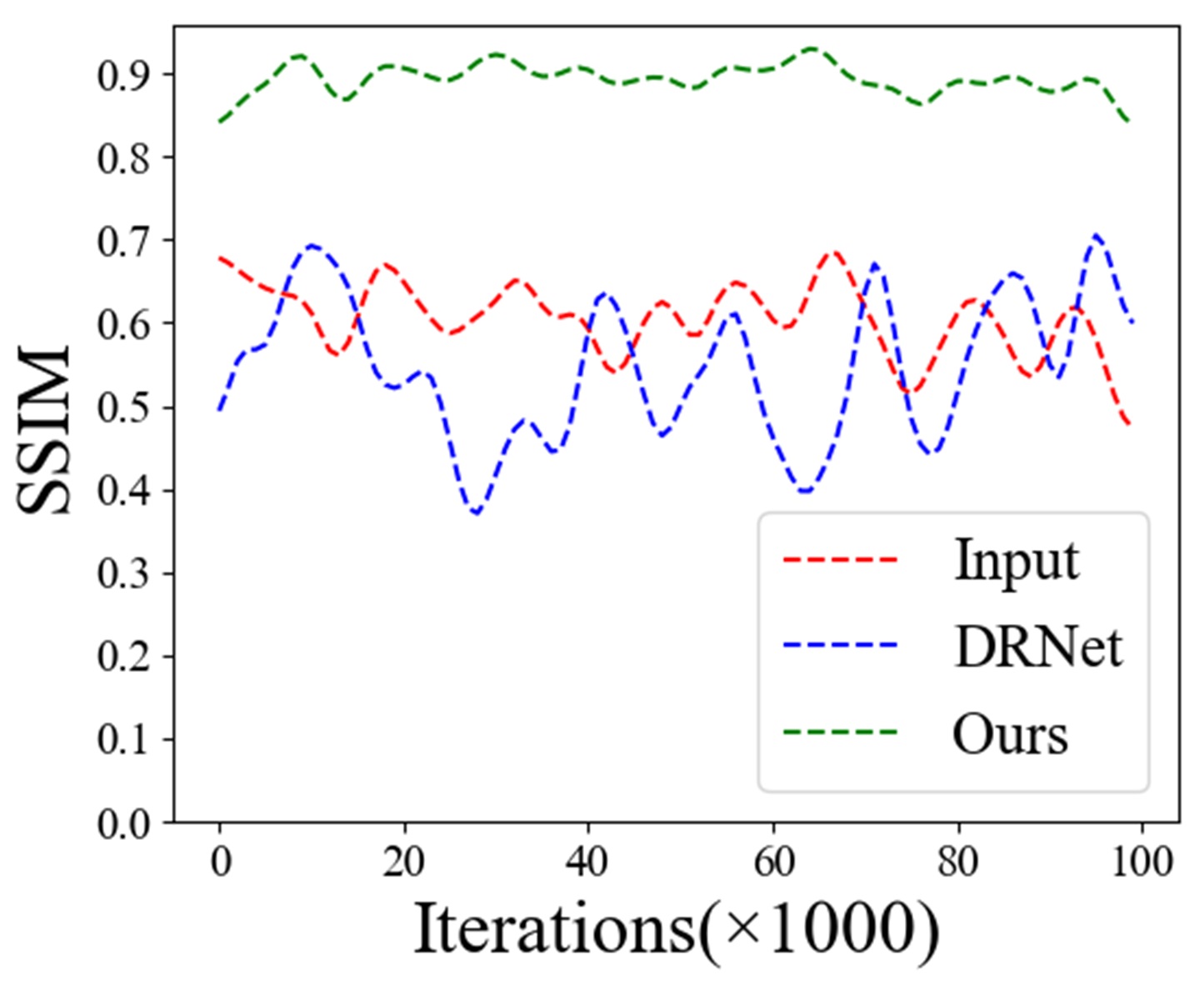}}
			\setlength{\abovecaptionskip}{-0.1cm}
			\setlength{\belowcaptionskip}{-0.1cm}
			\caption{Online training PSNR and SSIM during the 100k iterations.}
			\label{fig:8}
		\end{minipage}
\end{figure}

\subsection{Real Noise Removal}

X-ray computed tomography (CT) is widely used as important imaging modalities in modern clinical diagnosis. Considering the potential radiation risk to the patient, lowering the radiation dose increases the noise and artifacts in reconstructed images, which can compromise diagnostic information. Typically, noise in x-ray photon measurements can be simply modeled as the combination of Poisson quantum noise and Gaussian electronic noise. However, the noise in reconstructed images is more complicated and does not obey any statistical distribution across the whole image. Therefore, classical image post-processing methods based on noise statistic prior, \eg, {\itshape N2N}, are unsuitable for Low-dose CT (LDCT) denoising.

A real clinical dataset authorized by Mayo Clinic for {\itshape the 2016 NIH-AAPM-Mayo Clinic Low Dose CT Grand Challenge}\footnote {\url {http://www.aapm.org/GrandChallenge/LowDoseCT/}} is used to evaluate LDCT image reconstruction algorithms, which contains 5936 images in 512$\times$512 resolution from 10 different subjects. We randomly select 4000 images as training set, the remaining is as testing set. {\itshape DIP}, {\itshape BM3D} and {\itshape RedCNN} \cite{Chen2017LowDoseCW}, which is an extended version of {\itshape RedNet}, are selected for evaluation in our experiments. The representative results are shown in Fig. \ref{fig:6}, {\itshape BM3D} introduces waxy artifacts into the reconstructed image. {\itshape DIP} fails to generate the fine local structures. {\itshape RedCNN} tends to generate smoother images. Our approach achieves the better balance between visual quality and noise removal. Table. \ref{table:3} gives the quantitative results.

\subsection{Ablation Study}

In this section, we perform an ablation study to analyze the effects of discrete disentangling representation and self-supervised modules in the proposed framework. Both quantitative and qualitative results on Gaussian noise removal are shown for the following three variants of the proposed method where each component is separately studied: a) Remove the noise encoder branch; b) Remove the representation adversarial network $D^{\mathcal{R}}$, directly learn the representations $z_{\mathcal{X}}$ and $z_{\mathcal{Y}}$ by the target domain constraints only; c) Remove the background consistency constraint from self-supervised modules, only retain the semantic consistency constraints.

The representative results are shown in Fig. \ref{fig:7}. Compared with the full model, referred to as (d), directly learning invariant representations from noise images would lead to the generator producing over-smooth results for (a) due to unexpected noise contained in features, which requires a powerful domain generator. Although (b) gives the better PSNR and SSIM after removing the feature adversarial module, some undesired artifacts adhere to high-frequency signals. Due to failing to provide the effective self-supervised constraint for the recovered images, although retaining the semantic consistency module, the model (c) also produces domain shift problems in generated images, \eg, inconsistency brightness and blurred details, resulting in worse visual perception. Quantitative results are shown in Table. \ref{table:4}.

In addition, considering {\itshape DRNet} \cite{Lu2019UnsupervisedDD} has similar architecture with ours, which extends {\itshape DRIT} \cite{Lee2018DiverseIT} while introducing extra feature loss to solve image deblurring, we select it as a representative domain transfer method to compare the convergence of algorithms on denoising task. Fig. \ref{fig:8} gives the convergence plots for AWAN removal, where we trained two models from scratch on the same training set. Although {\itshape DRNet} also uses the similar idea of disentangled representation to solve image restoration, which is different from ours in essence.  Varying noise-levels and types lead to unstable learning during training due to lack of clear domain boundary. Aiming to learn invariant representation, our method gives faster and more stable convergence plots.
\section{Conclusion}

In this paper, we propose an unsupervised learning method for image restoration. Specifically, we aim to learn invariant representations from noise data via disentangling representations and adversarial domain adaption. Aided by effective self-supervised constraints, our method could reconstruct the higher-quality images with finer details and better visual perception. Experiments on synthetic and real image denoising show our method achieves comparable performance with other state-of-the-art methods, and has faster and more stable convergence than other domain adaption methods.

\section*{Acknowledge}

This work is supported by the National Natural Science Foundation of China under grant 61871277, and in part by the Science and Technology Project of Sichuan Province of China under grant 2019YFH0193.

{\small
\bibliographystyle{ieee_fullname}
\bibliography{egbib}
}

\end{document}